\pdfoutput=1

\documentclass[11pt,a4paper]{article}
\usepackage[acceptedWithA]{tacl2018v2}

\usepackage{microtype}

\usepackage{xspace,mfirstuc,tabulary}

\newif\iftaclinstructions
\taclinstructionsfalse %
\iftaclinstructions
\renewcommand{\confidential}{}
\renewcommand{\anonsubtext}{(No author info supplied here, for consistency with
TACL-submission anonymization requirements)}
\newcommand{\instr}
\fi

\iftaclpubformat %

\else

\fi

\usepackage{times}
\usepackage{latexsym}

\usepackage{amsmath}
\usepackage{stmaryrd}
\usepackage{amsfonts}
\usepackage{xspace}
\usepackage{nicefrac}
\usepackage{xfrac}
\usepackage{booktabs}
\usepackage{subcaption}
\usepackage{cancel}
\usepackage{makecell}
\usepackage{adjustbox}
\usepackage{color}
\usepackage{amsthm}
\usepackage{cleveref}
\usepackage{multirow}
\newtheorem{theorem}{Theorem}[section]
\newtheorem{lemma}[theorem]{Lemma}
\newtheorem{corollary}{Corollary}[theorem]

\usepackage{algorithm}
\usepackage{colortbl}
\usepackage{tabstackengine}
\usepackage{tikz}
\usepackage{relsize}
\usepackage{microtype}

\usepackage[noend]{algpseudocode}
\algnewcommand{\parState}[1]{\State%
    \parbox[t]{\dimexpr\linewidth-\algmargin}{\strut\hangindent=\algorithmicindent \hangafter=1 #1\strut}}

\algrenewcommand\algorithmicindent{1.0em}%

\theoremstyle{plain}

\theoremstyle{definition}
\newtheorem{definition}{Definition}[section]

\DeclareMathOperator*{\argmax}{argmax}

\definecolor{darkgrey}{rgb}{0.2,0.2,0.2}
\definecolor{grey}{rgb}{0.9,0.9,0.9}
\definecolor{darkblue}{rgb}{0.0,0.0,0.5}
\definecolor{darkred}{rgb}{0.5,0.0,0.0}
\definecolor{darkorange}{rgb}{1.0,0.55,0.0}
\definecolor{darkgreen}{rgb}{0.0,0.6,0.0}
\definecolor{darkyellow}{rgb}{1.0,0.65,0.0}
\definecolor{darkorange}{rgb}{1.0,0.65,0.0}
\definecolor{darkergreen}{rgb}{0.0,0.4,0.0}
\definecolor{lightblue}{rgb}{0.8,0.8,1.0}
\definecolor{lightgreen}{rgb}{0.8,1.0,0.8}
\definecolor{lightred}{rgb}{1.0,0.8,0.8}
\definecolor{lightyellow}{rgb}{1.0,1.0,0.8}
\definecolor{lightorange}{rgb}{1.0,0.9,0.8}
\definecolor{lightgrey}{rgb}{0.96,0.97,0.98}
\definecolor{brilliantlavender}{rgb}{0.96, 0.73, 1.0}

\definecolor{ryanred}{rgb}{0.64, 0.0, 0.0}
\definecolor{ryanblue}{rgb}{0.13, 0.0, 0.58}
\definecolor{ryangreen}{rgb}{0.12, 0.59, 0.0}
\definecolor{ryanpurple}{rgb}{0.65, 0.0, 0.57}

\newcommand*{\numberingBlueB}[1]{%
  \protect\tikz[baseline={([yshift=-1.5pt]n.base)}]%
  \protect\node[fill=blue!25,shape=circle,inner sep=1pt,draw](n){\small #1};}

\newcommand*{\numberingRedB}[1]{%
  \protect\tikz[baseline={([yshift=-1.5pt]n.base)}]%
  \protect\node[fill=red!15 ,shape=circle,inner sep=1pt,draw](n){\small #1};}

\newcommand*{\numberingGreenB}[1]{%
  \protect\tikz[baseline={([yshift=-1.5pt]n.base)}]%
  \protect\node[fill=green!25,shape=circle,inner sep=1pt,draw](n){\small #1};}

\newcommand*{\numberingYellowB}[1]{%
  \protect\tikz[baseline={([yshift=-1.5pt]n.base)}]%
  \protect\node[fill=yellow!25,shape=circle,inner sep=1pt,draw](n){\small #1};}

\crefname{section}{\S}{\S\S}
\crefname{table}{Tab.}{}
\crefname{figure}{Fig.}{}
\crefname{algorithm}{Alg.}{}
\crefname{appendix}{App.}{}
\crefname{lemma}{Lemma}{}
\crefname{theorem}{Theorem}{}
\crefname{prop}{Proposition}{}
\crefname{cor}{Corollary}{}
\crefname{align}{}{}
\crefname{equation}{}{}

\newcommand{\bleu}{\textsc{bleu}\xspace}

\newcommand{\xx}{\mathbf{x}}
\newcommand{\yy}{\mathbf{y}}
\newcommand{\bv}{\mathbf{v}}
\newcommand{\calYs}{\mathcal{Y}}

\newcommand{\calO}{\mathcal{O}}

\newcommand{\vocab}{\mathcal{V}}

\newcommand{\defeq}[0]{\overset{\smaller\mathrm{def}}{=}}

\newcommand{\calHbs}{\mathcal{H}_{\textsc{bs}}}
\newcommand{\calHdk}{\mathcal{H}_{\textsc{A}}}
\newcommand{\algname}{best-first beam search}
\newcommand{\Algname}{Best-first beam search}
\newcommand{\AlgName}{Best-First Beam Search}

\usepackage[hang,flushmargin]{footmisc}

\newcommand{\calQ}{\mathcal{Q}}

\newcommand{\eps}{\varepsilon}

\newcommand{\vp}{\mathbf{y}}
\newcommand{\nmax}{n_{\textit{max}}}
\newcommand{\score}{\mathrm{score}}
\newcommand{\scorestwos}{\score_{\textit{s2s}}}
\newcommand{\astar}{\text{A}^*\xspace}
\newcommand{\dd}[1]{\textcolor{darkgreen}{\bf\fontsize{7.5}{12} \selectfont \!(#1\%)}}
\newcommand{\db}[1]{\textcolor{darkgreen}{\bf\fontsize{7.5}{12} \selectfont \!+#1}}

\newcommand{\bos}{\textsc{bos}\xspace}
\newcommand{\eos}{\textsc{eos}\xspace}

\usepackage[disable]{todonotes}
\makeatletter
\newcommand*\iftodonotes{\if@todonotes@disabled\expandafter\@secondoftwo\else\expandafter\@firstoftwo\fi}  %
\makeatother

\newcommand{\note}[4][]{\todo[author=#2,color=#3,size=\scriptsize,fancyline,caption={},#1]{#4}} %
\newcommand{\ryan}[2][]{\note[#1]{ryan}{violet!40}{#2}}
\newcommand{\clara}[2][]{\note[#1]{clara}{orange}{#2}}

\newcommand{\timv}[2][]{\note[#1]{timv}{magenta!40}{#2}}

\usepackage{mathabx}
\newcommand{\ucambridge}{\Asterisk }
\newcommand{\ethz}{\bigstar}
\newcommand{\jhu}{\lightning}

\title{\AlgName{}}

\author{Clara Meister$^\ethz$~\;~Tim Vieira$^\jhu$~\;~Ryan Cotterell$^{\ucambridge,\ethz}$ \\
  $^\ethz$ETH Z\"{u}rich~\;~$^\jhu$Johns Hopkins University~\;~$^\ucambridge$University of Cambridge \\
  \texttt{clara.meister@inf.ethz.ch}~\;~\texttt{tim.vieira@gmail.com} \\ \texttt{ryan.cotterell@inf.ethz.ch}}

\date{}

\begin{document}
\setlength{\abovedisplayskip}{8pt}
\setlength{\belowdisplayskip}{8pt}
\setlength{\abovedisplayshortskip}{6pt}
\setlength{\belowdisplayshortskip}{6pt}
\maketitle

\begin{abstract}
Decoding for many NLP tasks requires an effective heuristic algorithm for approximating exact search since the problem of searching the full output space is often intractable, or impractical in many settings.  The default algorithm for this job is beam search---a pruned version of breadth-first search.  Quite surprisingly, beam search often returns \emph{better} results than exact inference due to \emph{beneficial} search bias for NLP tasks.
In this work, we show that the standard implementation of beam search
can be made up to 10x faster in practice.
Our method assumes that the scoring function is monotonic in the sequence length, 
which allows us to safely prune hypotheses that cannot be in the final set of hypotheses early on.
We devise effective monotonic approximations to popular nonmonontic scoring functions, including length normalization and mutual information decoding.  Lastly, we propose a memory-reduced variant of \algname{}, which has a similar beneficial search bias in terms of downstream performance, but runs in a fraction of the time.\looseness=-1
\end{abstract}

\section{Introduction}

Beam search is a common heuristic algorithm for decoding structured predictors, e.g., neural machine translation models and transition-based parsers. 
Due to the widespread adoption of recurrent neural networks and other non-Markov models, 
traditional dynamic programming solutions, such as the Viterbi algorithm \cite{viterbi1967error}, are prohibitively inefficient; this makes beam search a common component of many state-of-the-art NLP systems.  
Despite offering no formal guarantee of finding the highest-scoring hypothesis under the model, beam search yields impressive
performance on a variety of tasks---unexpectedly providing a beneficial search bias over \emph{exact} search for many tasks \cite{stahlberg-byrne-2019-nmt}.

Within NLP, most research on beam search has focused on altering the log-probability scoring function to return improved results, e.g., higher \bleu scores \cite{Wu2016GooglesNM, murray-chiang-2018-correcting, shu-nakayama-2018-improving, yang-etal-2018-breaking} or a more diverse set of outputs \cite{diverse-beam-search}.
However, little work has been done
to speed up beam search itself. Filling this gap, this paper focuses on reformulating beam search in order to make it \emph{faster}. 
We propose \algname, a prioritized version of traditional beam search which is up to an order of magnitude faster in practice while still returning the same set of results.  We additionally discuss an even faster heuristic version of our algorithm which further limits the number of candidate solutions, leading to a smaller memory footprint while still finding good solutions.

Concretely, we offer a novel interpretation of beam search as an agenda-based algorithm where
traditional beam search is recovered
by employing a length-based prioritization scheme. 
We prove that a specific best-first prioritization scheme, as in classic A$^*$ search \cite{hart1968formal}, allows for the elimination of paths that will necessarily fall off the beam; for many scoring functions, including standard log-probability scoring, we can still guarantee the same $k$ hypotheses as traditional beam search are returned. Indeed, our algorithm returns beam search's top hypothesis the first time it encounters a complete hypothesis, allowing the program to stop early. 
Further, we discuss the application of \algname{} to several popular scoring functions in the literature \cite{he-2016-length, li-etal-2016-diversity}; this demonstrates that we have a general framework for adapting a variety of rescoring methods and alternate objectives to work with our algorithm.\looseness=-1

Empirically, we compare \algname{} to ordinary beam search on two
NLP sequence-to-sequence tasks: neural machine translation (NMT) and abstractive summarization (AS). On NMT, we find that our algorithm achieves roughly a 30\%
speed-up over traditional beam search with increased gains for larger beams (e.g., $\approx{10}$x for a beam of 500). We find similar results hold for AS.
Finally, we show that our memory-reduced version, which limits the number of active hypotheses, leads to additional speed-ups over \algname{} across beam sizes while maintaining similar \bleu scores. Our code is available online at \url{https://github.com/rycolab/bfbs}\looseness=-1

\section{Sequence Transduction}
A core operation in structured prediction models is the determination of the highest-scoring output for a given input under a learned scoring model.\looseness=-1
\begin{equation}
    \yy^\star \defeq \argmax_{\yy \in \calYs(\xx)}\ \score(\xx, \yy)  \label{eq:general-argmax-problem}
\end{equation}
where  $\xx$ is an input and $\mathcal{Y}(\xx)$ is a set of well-formed outputs for the input.  An important example of \cref{eq:general-argmax-problem} is maximum a posteriori (MAP), 
\begin{equation}
\yy^{\mathrm{MAP}} \defeq \argmax_{\yy \in \mathcal{Y}(\xx)}\ p(\yy \mid \xx).
\label{eq:MAP}
\end{equation}

Our work focuses on sequence-to-sequence transduction: predicting an output sequence given an input sequence. One such task is machine translation, wherein a source-language sentence is mapped (``transduced'') to a target-language sentence.  While our exposition focuses on sequence-to-sequence prediction, our algorithms are directly applicable to any sequential structured prediction model, such as transition-based parsers \cite{nivre-etal-2008-parsing} and sequence taggers \cite{MEMM, laffertyCrf}.\looseness=-1

\paragraph{Notation.}
Let $\xx = \langle x_1, \ldots, x_{N_\xx} \rangle$ be an input sequence of length $N_{\xx}$ and, likewise, let $\yy =
\langle y_1, \ldots , y_{N_{\yy}} \rangle$ be an output sequence of length $N_{\yy}$. 
Each $y_t$ is an element of $\vocab$, the set of output tokens. \ryan{Are we consistent with input and output?}\clara{can't find any inconsistencies so far but keeping this as a reminder to do another sweep}
Finally, let $\calYs(\xx)$ be the set of all valid output
sequences (i.e., complete hypotheses). For the task of language generation, which we focus on experimentally, this set is defined as\looseness=-1
\begin{equation}
    \calYs(\xx) \defeq \{ \bos \circ \bv \circ \eos \mid \bv \in \vocab^{< \nmax} \}
    \label{eq:vocab}
\end{equation}
\noindent where $\circ$ is string concatenation and $\vocab^{< \nmax(\xx)}$ is the set of all subsets of $\vocab^\star$ of size $< \nmax(\xx)$.
In words, every valid sequence begins and ends with distinguished tokens (\bos and \eos, respectively).\footnote{\bos and \eos are typically members of $\vocab$. Often, \eos counts towards the $\nmax$ length limit while \bos does not. This is reflected in \cref{eq:vocab}.} Furthermore, each sequence has \emph{at most} length $\nmax(\xx)$---which is typically dependent on $\xx$---a restriction we impose to ensure termination.  Some applications may require a stronger coupling between $\calYs(\xx)$ and $\xx$ (e.g., $|\xx| = |\yy|$).  We drop the dependence of $\calYs$ and $\nmax$ on $\xx$ when it is clear from context.

\paragraph{Scoring.} We consider a general additively decomposable scoring model of the form
\begin{align}
    \score(\xx, \yy) &= \sum_{t=1}^{N_{\yy}} \score(\xx, \yy_{<t} \circ y_t) \label{eq:additive}
\end{align}
This framework covers a variety of modeling methodologies including probabilistic  transducers (both globally and locally normalized) and non-probabilistic models such as maximum-margin techniques \cite{taskar_mmmn}.  Most importantly, \cref{eq:additive} covers MAP decoding \cref{eq:MAP} of neural sequence-to-sequence models {\`a} la \newcite{sutskever-2014}:\footnote{To see why, apply $\exp$ (an order-preserving transformation):
$\exp (\score_{\textit{s2s}}(\xx, \yy)) = 
\exp\left(\sum_{t=1}^{N_{\yy}} \log p(y_t \mid \yy_{<t}, \xx) \right)  
= \prod_{t=1}^{N_{\yy}} p(y_t \mid \yy_{<t}, \xx)
= p(\yy \mid \xx)$.}%
\begin{equation}
    \scorestwos(\xx, \yy_{<t} \circ y_t) = \log p(y_t \mid \yy_{<t}, \xx)
    \label{eq:standard}
\end{equation}
\noindent We note that \cref{eq:standard} is the scoring function used for decoding many language generation models. \clara{citations}

\paragraph{Beam search.}

The worst-case running time of exactly computing \cref{eq:general-argmax-problem} is exponential in $\nmax$; namely, $\calO(|\vocab|^{\nmax})$.\footnote{This can be improved if, for example, $\score(\cdot, \cdot)$ admits a low-order Markov factorization \citep{viterbi1967error,vieira-etal-2016-speed2}.  We do not discuss that setting in this paper because it limits the scoring model's expressive power.
} 
Beam search is a commonly used approximation to \cref{eq:general-argmax-problem} in NMT and language generation tasks. It is used in many (if not most) state-of-the-art NLP systems \cite{Wu2016GooglesNM, AAAI1714571, edunov-etal-2018-understanding, XLNET}.  
Beam search may be understood as a pruned version of
the classic path-search algorithm, breadth-first search (BFS), where the breadth is narrowed to the beam size $k$.  Pseudocode is given in \cref{alg:beam}.

Although, beam search does not solve \cref{eq:general-argmax-problem} exactly, it is a surprisingly useful approximation for NLP models.  In many settings, beam search \emph{outperforms} exact methods in terms of downstream evaluation \cite{koehn-knowles-2017-six, stahlberg-byrne-2019-nmt}.  For the remainder of this paper, we will pivot our attention away from exact solutions to \cref{eq:general-argmax-problem} to exact solutions to the \emph{beam search} output.\looseness=-1

\begin{definition}{\textit{$k$-optimal hypothesis.}}
We say that a hypothesis is $k$-optimal if it is the top hypothesis returned by beam search with beam size $k$.
\end{definition}

\algrenewcommand{\algorithmiccomment}[1]{\hskip1em$\rightarrow$ \footnotesize#1 \normalsize}
\begin{algorithm}[tb]
\textbf{Input:} $\xx$: source sentence\\
\hspace*{2.7em} $k$: maximum beam size \\
\hspace*{2.7em} $\nmax$: maximum hypothesis length \\
\hspace*{2.7em} $\score(\cdot, \cdot)$: scoring function 
\begin{algorithmic}[1]
\State $B_0 \gets \{ \langle 0, \bos \rangle \}$
\For{ $t \in \{ 1, \dots, n_{max} \}$ }
    \State $B \gets \emptyset$
    \For{$ \langle s, \vp \rangle \in B_{t-1}$}    
    \If{$\vp.\mathrm{last}() = \eos$}
        \State $B.\mathrm{add}(\langle s, \vp  \rangle)$
        \State \textbf{continue}
    \EndIf
    \For{$y \in \vocab$}
        \State $s \gets \score(\xx, \vp \circ y)$ \label{line:bs-score-eval}
        \State $B.\mathrm{add}( \langle s, \vp \circ y \rangle)$
    \EndFor
    \EndFor
    \State $B_t \gets B.\mathrm{top}(k)$ 
    \EndFor
\State \Return $B_{n_{max}}.\mathrm{max}()$
\end{algorithmic}
\caption{Standard beam search\footnotemark}
\label{alg:beam}
\end{algorithm}
\addtocounter{footnote}{-1} %
\stepcounter{footnote}\footnotetext{Often, the $\score$ function is additively decomposable in $t$, such as \cref{eq:standard}. Implementations can exploit this fact to make each score evaluation (line \ref{line:bs-score-eval}) $\mathcal{O}(1)$ rather than $\mathcal{O}(t)$. We did not make this implementation detail explicit in \cref{alg:beam} or \cref{alg:general} for generality and simplicity.}

\section{A$^*$ Beam Search}
We develop a meta-algorithm that is parameterized by several choice points. Our general search algorithm for decoding (\cref{alg:general}) takes an arbitrary prioritization function, stopping criterion, and search heuristic. With certain values of these attributes, we recover many common search algorithms: greedy search, beam search, best-first search \cite{dijkstra1959note}, and A$^*$ search \cite{hart1968formal}. We propose an alternate prioritization function for beam search that allows for faster decoding while still returning the same $k$-optimal set of hypotheses.

\begin{algorithm}[tb]
\textbf{Input:} $\xx$: source sentence\\
\hspace*{2.7em} $\nmax$: maximum hypothesis length \\
\hspace*{2.7em} $\score(\cdot, \cdot)$: scoring function \\
\hspace*{2.7em } $\varogreaterthan$: comparator \numberingRedB{1}\\
\hspace*{2.7em } $\mathrm{stop}(\cdot)$: stopping criterion \numberingBlueB{2}\\
\hspace*{2.7em} $k$: maximum beam size \numberingGreenB{3} \\
\hspace*{2.7em} $h(\cdot, \cdot)$: heuristic function  \numberingYellowB{4} 
\begin{algorithmic}[1]
\State $\calQ \gets$\hspace*{-\fboxsep}\colorbox{red!15}{\parbox{85pt}{$\mathrm{priority\_queue}(\varogreaterthan)$}} 
\State $\calQ.\mathrm{push}(\langle 0, \bos\rangle)$
\State $\textsc{pops} \gets \mathrm{counter}()$
\While{\textbf{not}  \hspace*{-\fboxsep}\colorbox{blue!25}{\parbox{35pt}{$\mathrm{stop}(\calQ)$}} \textbf{and not} $\calQ.\mathrm{empty}()$} 
    \State $\langle s_h, \vp \rangle \gets \calQ.\mathrm{pop}()$ \newline
    \hspace*{-\fboxsep}\colorbox{green!25}{\parbox{\linewidth}{
      \If{$\textsc{pops}[|\vp|] \geq k$ \textbf{or} $|\vp| > \nmax$} \State \textbf{continue} 
      \EndIf
      \State $\textsc{pops}[|\vp|] \gets \textsc{pops}[|\vp|] + 1$ 
    }}
    \If{$\vp.\mathrm{last}() = \eos$}
        \State $\calQ.\mathrm{push}(\langle s_h, \vp \circ \eos \rangle)$
    \Else{:}
    \For{$y \in \vocab$}
        \State $s \gets  \score(\xx, \vp \circ y)$
        \State $s_h \gets s +$ \hspace*{-\fboxsep}\colorbox{yellow!25}{\parbox{50pt}{ $h(\xx, \vp \circ y)$}}
        \State $\calQ.\mathrm{push}(\langle s_h, \vp \circ y \rangle)$ \label{line:push}
    \EndFor
    \EndIf  
   
\EndWhile
\State \Return $\calQ.\mathrm{pop}()$ \textbf{if not} $\calQ.\mathrm{empty}()$ \textbf{else null}

\end{algorithmic}
\caption{General decoding scheme.\protect\footnotemark[\value{footnote}]$^,$\footnotemark{} Highlighted sections are choice points in the algorithm for which values determine the search strategy. See \cref{sec:choice_points} for detailed explanation. }
\label{alg:general}
\end{algorithm}
\footnotetext{If the last token of $\yy'$ is the end symbol (e.g., \eos), then $\yy'$ is not expanded any further. One can either regard $\yy'$ as any other hypothesis albeit with $\yy' \circ y_t = \yy'$ or keep appending \eos{} (i.e  $\yy' \circ y_t = \yy' \circ \eos$ ) so that time step and length can be regarded as synonymous. We adopt the latter standard for comparability with subsequent algorithms.}

\subsection{Choice Points of \cref{alg:general}}\label{sec:choice_points}
Here we review the components of our meta algorithm (the highlighted sections in \cref{alg:general}) that can be varied to recover different search strategies:
\begin{itemize}
 \item[\numberingRedB{1}] $\varogreaterthan: \yy\times\yy \to \{\text{True, False}\}$.\clara{doing binary T/F felt a bit strange} A priority queue $\calQ$ maintains the set of active hypotheses.  Elements in this set are ordered according to a generic comparator $\varogreaterthan$.  When its $\mathrm{peek}()$ (or $\mathrm{pop}()$) methods are called, the first element ordered by $\varogreaterthan$ is returned (or returned and removed).\looseness=-1
\item[\numberingBlueB{2}] $\mathrm{stop}(\cdot): \mathrm{Collection}\langle \yy\rangle \to \{\text{True, False}\}$. The algorithm terminates according to configurable stopping criterion based on the current set of elements in $\calQ$.
\item[\numberingGreenB{3}] $k \in \mathbb{N}_{> 0}$. Only $k$ paths of a given length are considered. If the algorithm has already encountered $k$ paths of a given length, subsequent paths of that length are not evaluated. If we take $k=\infty$, we recover
unpruned search algorithms.\looseness=-1
\item[\numberingYellowB{4}] $h(\cdot,\cdot): \xx \times \yy \to \mathbb{R}$. A heuristic function $h(\xx,\yy)$ can be used during search to change the priority in which paths are evaluated. We note that with pruning, a heuristic may change the value of the $k$-optimal hypothesis (see \cref{sec:correct}). 
\end{itemize}

\begin{table*}
\fontsize{10}{12} \selectfont
\centering

\begin{adjustbox}{width=2\columnwidth}
\begin{tabular}{ccccc}
\toprule
&
\textbf{Beam Search} &
\textbf{Best-First Beam Search} &
\textbf{A$^*$ Beam Search} \\
\midrule
\rowcolor{red!15}\numberingRedB{1} &
     \tabularCenterstack{c}{$\langle s_h, \yy \rangle \varogreaterthan \langle s_h', \yy' \rangle \iff |\yy| < |\yy|' $ \\
     $  \textbf{ or } \left(|\yy| = |\yy|' \textbf{ and } s_h \geq s_h' \right)$} & 
      \tabularCenterstack{c}{$\langle s_h, \yy \rangle \varogreaterthan \langle s_h', \yy' \rangle \iff s_h > s_h'$ \\ $  \textbf{ or } \left(s_h = s_h' \textbf{ and } |\yy| < |\yy|' \right)$}& 
      \tabularCenterstack{c}{$\langle s_h, \yy \rangle \varogreaterthan \langle s_h', \yy' \rangle \iff s_h > s_h' $ \\ $\textbf{ or } \left(s_h = s_h' \textbf{ and } |\yy| < |\yy|' \right)$} \\
\rowcolor{blue!25}\numberingBlueB{2} & \tabularCenterstack{c}{$\mathrm{stop}{(\calQ)} \iff$\\ $\yy.\mathrm{last}() = \eos \quad \forall \yy \in \calQ$ }& \tabularCenterstack{c}{$\mathrm{stop}(\calQ) \iff$\\ $\calQ.\mathrm{peek}().\mathrm{last}() = \eos$ }& \tabularCenterstack{c}{$\mathrm{stop}(\calQ) \iff$\\ $\calQ.\mathrm{peek}().\mathrm{last}() = \eos$ }\\
\rowcolor{green!25}\numberingGreenB{3} & $k = $ beam size  & $k=$ beam size& $k = $ beam size\\
\rowcolor{yellow!25}\numberingYellowB{4} &0 & 0 & any admissible heuristic \\
\toprule
&
\textbf{Breadth-First Search} &
\textbf{Best-First Search} &
\textbf{A$^*$ Search} \\ \midrule
\rowcolor{red!15}\numberingRedB{1} &
      \tabularCenterstack{c}{$\langle s_h, \yy \rangle \varogreaterthan \langle s_h', \yy' \rangle \iff |\yy| < |\yy|' $ \\
     $  \textbf{ or } \left(|\yy| = |\yy|' \textbf{ and } s_h \geq s_h' \right)$} & 
      \tabularCenterstack{c}{$\langle s_h, \yy \rangle \varogreaterthan \langle s_h', \yy' \rangle \iff s_h > s_h'$ \\ $  \textbf{ or } \left(s_h = s_h' \textbf{ and } |\yy| < |\yy|' \right)$}& 
      \tabularCenterstack{c}{$\langle s_h, \yy \rangle \varogreaterthan \langle s_h', \yy' \rangle \iff s_h > s_h' $ \\ $\textbf{ or } \left(s_h = s_h' \textbf{ and } |\yy| < |\yy|' \right)$} \\
\rowcolor{blue!25}\numberingBlueB{2} & \tabularCenterstack{c}{$\mathrm{stop}{(\calQ)} \iff$\\ $\yy.\mathrm{last}() = \eos \quad \forall \yy \in \calQ$ } & \tabularCenterstack{c}{$\mathrm{stop}{(\calQ)} \iff$\\ $\calQ.\mathrm{peek}().\mathrm{last}() = \eos$ }& \tabularCenterstack{c}{$\mathrm{stop}{(\calQ)} \iff$\\ $\calQ.\mathrm{peek}().\mathrm{last}() = \eos$ }\\
\rowcolor{green!25}\numberingGreenB{3} & $k = \infty$  & $k=\infty$ & $k = \infty$ \\
\rowcolor{yellow!25}\numberingYellowB{4} &0 & 0 & any admissible heuristic  \\
\bottomrule
\end{tabular}
\end{adjustbox}
\caption{Values at choice points for various search algorithms. Note that any admissible heuristic may be used for variants of $\astar$ search.} 
\label{tbl:choice_points}
\end{table*}

\paragraph{Recovering Beam Search.}
To recover beam search from \cref{alg:general}, we use the choice points from \cref{tbl:choice_points}. Explicitly, the comparator prioritizes hypotheses from earlier time steps first, but breaks ties with the hypotheses' scores under the model. We note that while the standard algorithm for beam search does not prioritize by score within a time step, variations of the algorithm use this strategy so they can employ early-stopping strategies \cite{klein-etal-2017-opennmt, huang-etal-2017-finish}. 
Beam search terminates once either 
\emph{all} hypotheses end in \eos 
or the queue is empty (i.e., when the $k$ beams have been extended $\nmax$ time steps but none end in \eos). In the second case, no complete hypothesis is found. 
Finally, choosing the heuristic $h(\xx, \vp) = 0$ makes the algorithm a case of standard best-first search.

Note that, while standard beam search returns a set, \cref{alg:general} only returns the $k$-optimal hypothesis. This behavior is sufficient for the majority of use cases for beam search. However, if the full set of $k$ hypotheses is desired, the stopping criterion can be changed to evaluate true only when $k$ hypotheses are complete. Under the other beam search settings, this would provably return the same set as beam search (see \cref{sec:correct}).

\paragraph{Recovering A$^*$.}
To recover the traditional A$^*$ search algorithm, we use the comparator that prioritizes hypotheses with a higher score first; ties are broken by hypothesis length. The algorithm terminates when the first item of $\calQ$ contains an \eos. If we take $k=\infty$, \algname{} recovers A$^*$.  Any admissible heuristic may be used for $h(\xx, \vp)$.
\begin{definition}{\emph{Admissible Heuristic.}} A heuristic $h$ is admissible if it never overestimates the future cost---or underestimates the future reward---of continuing down a path.
\end{definition}

\subsection{\AlgName{}}
In its original form, $\astar$ search may traverse the entire $\calO(|\vocab|^{\nmax})$ graph, which as discussed earlier, is intractable for many decoding problems.
While standard beam search addresses this problem by limiting the search space, it still has computational inefficiencies---namely, we \emph{must} analyze $k$ hypotheses of a given length (i.e., time step), regardless of how poor their scores may already be, before considering longer hypotheses.
However, prioritization by length is not strictly necessary for finding a $k$-optimal hypothesis. 
As is done in $\astar$, we can use score as the prioritization scheme and still guarantee optimality--or $k$-optimality--of the paths returned by the algorithm.

We define $\astar$ beam search as the $\astar$ algorithm where breadth is limited to size $k$. Further, we define \algname{} as the case of $\astar$ beam search when no heuristic is used (see \cref{tbl:choice_points} for algorithm settings). This formulation has two large advantages over standard beam search: (1) we gain the ability to remove paths from the queue that are guaranteed to fall off the beam and (2) we can terminate the algorithm the first time a complete hypothesis is encountered.  We can therefore reduce the computation required for decoding while still returning the same set of results. 

The mathematical property that makes this short-circuiting of computation possible is the monotonicity of the scoring function.  Note that not all scoring functions are monotonic, but many important ones are, including log-probability \cref{eq:standard}.  We discuss effective approximations for popular non-monotonic scoring functions in \cref{sec:score}.
\begin{definition}{\emph{Monotonicity.}}\label{def:monotonic}
A scoring function $\score(\cdot, \cdot)$ is monotonic in $t$
if for all $\xx$, $\yy_{<t} = \langle y_1 \ldots y_{t-1} \rangle$, $y_t \in \vocab, \, 1 \le t \le \nmax$
\begin{align}
\score(\xx, \yy_{<t}) \ge \score(\xx,& \yy_{<t} \circ y_t)  \nonumber
\end{align}
\end{definition}
\noindent Clearly, \cref{eq:standard} is a monotonic scoring function in $t$ because $\scorestwos \leq 0$, that is, the $\score$ of a partial hypothesis $\yy_{<t}$ can only decrease if we extend it by another symbol $y_t$\clara{flesh out a bit more}. This implies we can order our search according to $\score(\xx,\yy_{<t})$ without fear of overlooking a hypothesis whose score would increase over time. Furthermore, once $k$ hypotheses of a given length $t$ have been evaluated, we no longer need to consider any hypothesis where $|\yy| < t$ since such hypotheses would necessarily fall off the beam. We can therefore remove such hypotheses from the queue and avoid wasting computational power on their evaluation. We prove this formally in \cref{sec:correct}.

Another implication of the monotonicity property of $\score$ is that we may terminate best-first beam search once a hypothesis containing $\eos$ is encountered (i.e., the end state is found). If the full set of $k$ complete hypotheses is desired, then we simply continue until $k$ hypotheses have reached \eos. We prove the $k$-optimality of these hypotheses under best-first beam search in \cref{sec:correct}.

\subsection{Implementation Details}\label{sec:imp} 
Standard beam search forms a separate set of active hypotheses for each time step, i.e., each $B_t$ is its own set. Once $B_t$ has been narrowed down to the top $k$, the previous $B_{<t}$ can be forgotten. However in \algname, since hypotheses are not evaluated in order of time step, we may need to keep $B_t$ from several time steps at any given point. 

A naive implementation of best-first beam search is to keep a single priority queue with all the active hypotheses ordered by current score. However, each push to the queue would then require $\calO(\log (\nmax k |\vocab|))$ time.
We can reduce this runtime by instead keeping a priority queue of beams, where the priority queue is ordered by the highest-scoring hypothesis from each beam. Further, each beam can be represented by a min-max queue \cite{min_max}; this allows us to limit the size of $B_t$ to $k$: we can check in $\calO(1)$ time if a hypothesis is in the top-$k$ before adding it to $B_t$.

A potential inefficiency, which we avoid, comes from updating $B_{t+1}$, which we must do when evaluating a hypothesis from $B_{t}$. Since all beams are stored in a queue, there is no guarantee of the location in the queue of $B_{t+1}$. To avoid $\calO(\nmax)$ lookup, we can keep a pointer to each beam, indexed by $t$ making the lookup $\calO(1)$. However, we acquire a $\calO(\log \nmax)$ term to update the queue of beams as $B_{t+1}$ may change priority.
\paragraph{Memory-Reduced \AlgName{}.}
A major drawback of the $\astar$ algorithm is its memory usage, which in the worst-case is $\calO(b^d)$ for breadth width $b$ and maximum depth $d$. In the $\astar$ formulation of beam search, where the breadth width is limited to the beam size, this amounts to worst-case $\calO(k \cdot {n_{max}})$ memory usage, where standard beam search has $\calO(k)$ memory usage. While in many settings the multiplicative factor may be insignificant, for neural sequence models it can be prohibitive; this is due to the large amount of memory required to store each hypothesis (e.g., prior hidden states needed to compute subsequent scores for scoring functions parameterized by neural networks).

We propose a variant of \algname{} that limits memory usage, i.e., the queue capacity. Specifically, if we reach the chosen queue capacity, we remove the worst scoring active hypothesis from the earliest active time step. This can easily be done in $\calO(1)$ time given our pointer to each beam.

\section{Algorithm Analysis}

\subsection{Correctness}\label{sec:correct}
We show the equivalence of the top hypothesis\footnote{\Algname{} is guaranteed to return the same set of $k$ hypotheses as beam search. We include the proof for only the top hypothesis for simplicity. The proof for set equality follows naturally.} returned by beam search and \algname{} when
$\score(\cdot, \cdot)$ is monotonically decreasing in $t$, length-based prioritization is used and the beam size $k$ is the same for both algorithms. Without loss of generality, we hold $\xx$ constant in all the following proofs.

Note that we take the terms \emph{pop} and \emph{push} from queue terminology. Specifically, ``popping a hypothesis'' refers to making it past line 7 of \cref{alg:general}, where  a hypothesis $\yy$ is expanded by $y_t \in \vocab$. In path search terminology, this would be equivalent to visiting a node and adding the edges from that node as potential paths to explore. Lastly, we refer to the priority queue used by beam search and  \algname{} as $\calQ_{\mathrm{BS}}$ and $\calQ_{\mathrm{A}^*}$, respectively.\looseness=-1
\begin{lemma}
\Algname{} evaluates all hypotheses of a given length $t$ in order of their score. 
\label{lem:order}
\end{lemma}

\begin{proof}
We prove the lemma by induction. The lemma holds trivially for the base case of hypotheses of length 0 because the only hypothesis of length 0 is $\langle\bos\rangle$.

Now, by the inductive hypothesis, suppose \cref{lem:order} holds for all hypotheses of length $< t$. We will show it
must also hold for hypotheses of length $t$.
Consider two competing hypotheses: $\yy = \yy_{<t} \circ y_t$ and $\yy' = \yy_{<t}' \circ y'_t$. Note that $|\yy_{<t}| = |\yy_{<t}'| = t-1$. Suppose $\score(\xx, \yy') < \score(\xx, \yy)$. 

\emph{Case 1}: $\score(\xx, \yy_{<t}') < \score(\xx, \yy_{<t})$. Then by induction, $\yy_{<t}$ popped first and $\yy$ is pushed to $\calQ$ before $\yy'$. Since $\score(\xx, \yy') < \score(\xx, \yy)$, $\yy$ will be popped before $\yy'$.

\emph{Case 2}: $\score(\xx, \yy_{<t}) < \score(\xx, \yy_{<t}')$. Then by induction, $\yy_{<t}'$ is popped first and $\yy'$ is added to $\calQ$ before $\yy$. But, since $\score(\xx, \yy’) < \score(\xx, \yy) \leq \score(\xx, \yy_{<t})$ by monotonicity, then $\yy_{<t}$ will be popped before $\yy’$. Consequently, $\yy$ will be pushed to $\calQ$ before $\yy'$ is evaluated. By the rules of the priority queue $\yy$ will be evaluated before $\yy'$. 

\emph{Case 3}: $\score(\xx, \yy’) = \score(\xx, \yy$).
The lemma holds if either $\yy$ or $\yy'$ is popped first.

By the principle of induction, \cref{lem:order} holds for all $t\in \mathbb{N}_{>0}$.
\end{proof}

\begin{lemma}
The first hypothesis that \algname{} pops that ends in \eos is k-optimal.
\label{lem:eos}
\end{lemma}

\begin{proof}
Let $\yy$ be the first hypothesis popped by \algname{} ending in \eos. By rules of the priority queue, no other active hypothesis has a higher score than $\yy$. Additionally, by monotonicity of the scoring function, no other hypothesis can subsequently have score greater than $\yy$. Therefore $\yy$ must be k-optimal.
\end{proof}
\begin{lemma}
If \algname{} pops a hypothesis, then beam search necessarily pops that same hypothesis.
\label{lem:evaluate}
\end{lemma} 

\begin{proof}
We prove the lemma by induction on hypothesis length. The base case holds trivially: For hypotheses of length $0$, both \algname{} and beam search must pop the $\langle\bos\rangle$ as it is the only item in the queue after initialization.

By the inductive hypothesis, suppose \cref{lem:evaluate} holds for hypotheses of length $< t$. Suppose \algname{} pops a hypothesis $\yy = \yy_{<t} \circ y_t$ of length $t$. 

\textit{Case 1:} \Algname{} pops $k$ hypotheses of length $t-1$ before popping $\yy$, which is of length $t$. The sets of hypotheses of length $t-1$ that each algorithm pops are necessarily the same by the inductive hypothesis and the fact that they have the same cardinality. If \algname{} pops $\yy$, which is of length $t$, then it must be in the top-$k$ highest-scoring hypotheses of length $t$ in $\calQ_{\mathrm{A}^*}$ by the rules of the priority queue. Consequently, it must be in the top-$k$ in $\calQ_{\mathrm{BS}}$. 

\textit{Case 2:} \Algname{} has popped fewer than $k$ hypotheses of length $t-1$ before popping $\yy$. Then, all remaining hypotheses of length $t-1$ in $\calQ_{\mathrm{A}^*}$ must have $\score(\xx, \yy'_{<t}) < \score (\xx, \yy)$ by the rules of the priority queue. By the monotonicity of the score function, all extensions of those $\yy'_{<t}$ will also have $\score(\xx, \yy'_{<t} \circ y'_{t}) < \score(\xx, \yy)$. Because none of $\yy_{<t}’ \circ y_{t}'$ has greater score than $\yy$, $\yy$ must be in $B_t$.
\end{proof}

\begin{corollary}
\Algname{} will never pop more hypotheses than beam search.
\end{corollary}

\begin{theorem}\ryan{Need to think and revisit}
Once \algname{} has popped $k$ hypotheses of length $t$, hypotheses from time steps $<t$ do not need to be popped.
\label{thm:prune}
\end{theorem}

\begin{proof}
This follows from \cref{lem:order}. If $k$ hypotheses of length $t$ have been popped, then these must be the top-$k$ hypotheses from time step $t$. Therefore no hypothesis from time step $<t$ that is still in $\calQ_{\mathrm{A}^*}$ would be in the top-$k$ at time step $t$.
\end{proof}

\begin{theorem}
Let $\calHbs$ and $\calHdk$ be the set of $k$ hypotheses returned by beam search and \algname, respectively. $\calHbs = \calHdk$. 
\end{theorem}

\begin{proof}
Since $|\calHbs| = |\calHdk| = k$, we only need to show $\yy \in \calHbs \Longrightarrow \yy \in \calHdk$. 

Suppose, by way of contradiction, there exists
a hypothesis $\yy \in \calHbs$ such that $\yy \not \in \calHdk$.
If $\yy \not \in \calHdk$ then we must not pop the prefix $\yy_{<t}$ (where $\yy = \yy_{<t} \circ \yy_{t:|\yy|}$) for some time step $t < |\yy|$. 

\emph{Case 1}: At some time step $t+j$ ($j \geq 0$), we pop $k$ partial hypotheses $\{\yy_{\leq t+j}^{(1)}, \dots, \yy_{\leq t+j}^{(k)}\}$ where $\yy_{\leq t+j} \not \in \{\yy_{\leq t+j}^{(1)}, \dots, \yy_{\leq t+j}^{(k)}\}$. By \cref{lem:order}, it must be that $\score(\xx, \yy_{\leq t+j}^{(i)}) > \score(\xx, \yy_{\leq t+j}$) $\forall i \in 1, \dots, k$.
This implies that for beam search, $\yy_{\leq t+j}$ would not be in the top-$k$ paths at time step $t+j$ since by \cref{lem:evaluate}, paths $\{\yy_{\leq t+j}^{(1)}, \dots, \yy_{\leq t+j}^{(k)}\}$ would also be evaluated by beam search. Therefore $\yy$ cannot be in $\calHbs$, which is a contradiction.

\emph{Case 2}: For no time step $t+j$ ($j \geq 0$) do we pop $k$ paths. This can only happen if the algorithm stops early, i.e we have found $k$ complete hypotheses $\yy^{(1)}, \dots, \yy^{(k)}$. If this is the case, then by rules of the priority queue, each $\yy^{(1)}, \dots, \yy^{(k)}$ must have score greater than $\score(\xx, \yy_{< t})$. By monotonicity of the score function, $\score(\xx, \yy^{(i)}) > \score(\xx, \yy$). This implies $\yy$ cannot be in $\calHbs$, which is a contradiction.\looseness=-1
\label{thm:same}
\end{proof}

\paragraph{Non-monotonic Scoring Functions.} 
Non-monotonic scoring functions (\cref{def:monotonic}) break the assumptions of \cref{thm:same}, in which case \algname{} is not guaranteed to return a $k$-optimal hypothesis. However, when the scoring function is boundable from above, we can alter the original stopping criterion (\numberingBlueB{2} in \cref{alg:general}) such that $k$-optimality is again guaranteed. 

Given our assumed restriction on the search space---namely, $|\yy^\star \in \calYs(\xx)| \leq \nmax(\xx)$---we can upper-bound the maximal score of any hypothesis under the scoring function in use. Formally, for any function $\score$ we have:
\begin{align}
\mathrm{stop}{(\calQ)} \iff & \nonumber \\
    \score(\xx,\hat{\yy}) \geq\ & \score(\xx, \yy') + \mathcal{U}(\xx,\yy') \nonumber \\
    &\qquad\qquad\quad \forall \yy' \in \calQ
\label{eq:stop}
\end{align}
\noindent where $\hat{\yy}$ is the best complete hypothesis found so far and $\mathcal{U}(\xx,\yy')$ is the $\score$ function-dependent upper bound on how much the score of $\yy'$ can increase as $\yy'$ is expanded further.\footnote{For monotonic scoring functions, we have $\mathcal{U}(\xx,\yy') = 0$.} In this situation, \algname{} only terminates once no other hypothesis in $\calQ$ can have a score greater than the best finished hypothesis. We note that \citet{huang-etal-2017-finish} use a similar scheme for optimal stopping with bounded length normalization. We discuss examples of non-monotonic scoring functions in \cref{sec:score}.\looseness=-1

\paragraph{A Note on Heuristics.} 
Our analysis shows the equivalence of beam search and \algname, i.e., when $h(\xx, \yy) = 0$. The analysis does \emph{not} hold for arbitrary admissible heuristics.  A poor heuristic, e.g., one that grossly overestimates the future score of continuing down one path, may cause other items to be pruned from \algname{} that otherwise would have remained on the beam in standard beam search.

\subsection{Runtime}

\begin{theorem}\label{thm:runtime}
The runtime of \algname{} is $\calO(\nmax k \, (|\vocab|\log(k) +  \log(\nmax)))$
\end{theorem}

\begin{proof}
We pop at most $\nmax \cdot k$ items. Each pop
requires us to push $|\vocab|$ items. Each push requires $\log(k)$ time when the priority queue is implemented with a min--max heap \cite{min_max} and incrementally pruned so that it has no more than $k$ items. After pushing those $|\vocab|$ items, we have to perform a percolation in the priority queue of priority queues which requiers $\log(\nmax)$ time. This yields 
$\calO(\nmax k \, (|\vocab|\log(k) +  \log(\nmax)))$ time.
\end{proof}

\begin{theorem}
The runtime of standard beam search is $\calO(\nmax\, k\, |\vocab| \log(k))$. 
\end{theorem}
\begin{proof}
The proof is the same as \cref{thm:runtime}, but we can forgo the percolation step in the queue of queues because standard beam search proceeds in order of hypothesis length. This yields $\calO(\nmax k|\vocab|\log(k))$. 
\end{proof}

While the theoretical bound of \algname{} has an additional log factor compared to standard beam search, we find this to be negligible in practice. Rather, we find number of calls to $\score$, the scoring function under our model (e.g., a neural network), is often the bottleneck operation when decoding neural networks (see \cref{sec:exps} for empirical evidence). 
In terms of this metric, the beam search algorithm makes $\calO(k\nmax)$ calls to $\score$, as $\score$ is called once for each active hypothesis in $B$ and $B$ may evolve for $\nmax$ rounds. The worst-case number of calls to $\score$ will be the same as for beam search, which follows from \cref{lem:evaluate}.

\section{Scoring Functions}\label{sec:score}
Even before the findings of \newcite{stahlberg-byrne-2019-nmt}, it was well known that the best-scoring hypothesis with respect to the traditional likelihood objective can be far from ideal in practice \cite{Wu2016GooglesNM, murray-chiang-2018-correcting, yang-etal-2018-breaking}. For language generation tasks specifically, the results returned by neural models using the standard scoring function are often short and default to high-frequency words \cite{Vinyals2015ANC, shen-etal-2016-minimum}. 

To alleviate such problems, methods that revise hypothesis scores to incorporate preferences for longer, less repetitive, or more diverse options have been introduced and are often used in practice. While most such techniques change the scoring function such that it is no longer monotonic, we can still guarantee the $k$-optimality of the returned hypothesis for (upper) bounded scoring functions using the methods discussed in \cref{sec:correct}.
In the remainder of this section, we present alternate scoring schemes adapted to work with \algname. Additionally, we present several heuristics which, while breaking the $k$-optimality guarantee, provide another set of decoding strategies worth exploring.\looseness=-1

\paragraph{Length Normalization.}\label{sec:length}
Length normalization is a widely-used hypothesis scoring method that aims to counteract the propensity for shorter sequences to have higher scores under neural models; this is done by normalizing scores by hypothesis length (see \citet{murray-chiang-2018-correcting} for more detail).

For early stopping in beam search with length normalization, \citet{huang-etal-2017-finish} propose bounding the additive length reward as the minimum of a pre-determined optimal sequence length ratio $r$ and the final sequence length $N_{\yy}$:
\begin{equation}
\begin{split}
    \score_{\textsc{ln}}(\xx, \yy) =\, &\score(\xx, \yy) \\ &\,\,+\,\beta\cdot \min\{r|\xx|, N_{\yy}\}
\end{split} \label{eq:norm_score1}
\end{equation}
\noindent where $\beta$ is the scaling parameter for the reward.
We note, however, that the same can be done with the maximum sequence length $n_{max}$ such that the traditional length reward used by \citet{he-2016-length} is recovered:
\begin{align}
    \score_{\textsc{ln}}(\xx, \yy) &= \score(\xx, \yy) + \beta \min\{n_{max}, N_{\yy}\} \nonumber \\
         &= \score(\xx, \yy) + \beta N_{\yy}
\label{eq:norm_score2}
\end{align}

We formally propose two methods for length normalization. We use the scoring functions in \cref{eq:norm_score1} or \cref{eq:norm_score2} with either: (1) the following heuristic:
\small
\begin{equation}
    h(\xx, \yy) = 
 \begin{cases}
    0 & \text{for } \yy.\mathrm{last}() = \eos \\
    \beta \max\{b - |\yy|, 0\} & \text{for } \yy.\mathrm{last}() \neq \eos
  \end{cases}
  \label{eq:norm_heur}
\end{equation}
\normalsize
\noindent where $b$ can be $r|\xx|$ or $\nmax$;\footnote{We enforce $r|\xx|< \nmax$.} or (2) stopping criterion as in \cref{eq:stop} albeit with scoring function $\score_{\textsc{ln}}$ and upper-bound function:\looseness=-1
\begin{equation}
    \mathcal{U}(\xx,\yy) = \beta\max\{0, b - |\yy|\}
\end{equation}
Despite their similarities, these two methods are not guaranteed to return the same results. While the second method will return the same $k$-optimal hypotheses as beam search, using a heuristic during pruned search means we can no longer guarantee the $k$-optimality of the results with respect to the scoring function as the heuristic may push hypotheses off of the beam.
We present experimental results for both methods in \cref{sec:exps}.

 \begin{table*}[!h]
  \centering
  \setlength\tabcolsep{3pt}
  \adjustbox{max width=\textwidth}{
  \begin{tabular}{ @{}llllllllllll@{} }
   \toprule
      & \multicolumn{4}{c}{\bf IWSLT'14 De-En} & \multicolumn{3}{c}{\bf MTTT Fr-En} & 
      \multicolumn{3}{c}{\bf CNN-DailyMail}\\
     & $k\!=\!5$ & $k\!=\!10$ & $k\!=\!100$ & $k\!=\!500$\,\,  
     & $k\!=\!10$ & $k\!=\!100$ & $k\!=\!500$\,\, 
     & $k\!=\!5$ & $k\!=\!10$ & $k\!=\!100$  \\
     & (\textcolor{darkblue}{\bf35.6})& (\textcolor{darkblue}{\bf35.4}) & (\textcolor{darkblue}{\bf34.7}) & (\textcolor{darkblue}{\bf7.9}) & (\textcolor{darkblue}{\bf33.0}) & (\textcolor{darkblue}{\bf9.9}) & (\textcolor{darkblue}{\bf1.2}) & (\textcolor{darkblue}{\bf31.5}) & (\textcolor{darkblue}{\bf30.9}) & (\textcolor{darkblue}{\bf29.1}) \\
    \hline
    BF beam search   & 93\dd{24} & 169\dd{36}& 1275\dd{79}& 1168\dd{736}
    &  184\dd{16} & 867\dd{138} & 885\dd{836}
    & 200\dd{33} & 305\dd{43}&  2960\dd{92}\\
    
    Beam search (ES)\, & 107\dd{7} & 210\dd{9}  & 2047\dd{12} &7685\dd{27}
     &196\dd{9}  & 1310\dd{58}& 4182\dd{98}
    & 224\dd{19}& 357\dd{22} &  3942\dd{59}\\
   Beam search & 115 & 229 & 2286 &9770
    & 214 & 2066 & 8281
    & 266 & 435 & 5673 \\
    
    \bottomrule
  \end{tabular} }
  \caption{Average number of calls (rounded to nearest whole digit) to $\score$, the sequence transduction model, per generated sequence when using different decoding algorithms. Green percentages are performance improvements over standard beam search. Beam search (ES) refers to the OpenNMT early-stopping method \cite{klein-etal-2017-opennmt}. All methods provably return the same solution and thus, evaluation metrics (in dark blue) for a given beam size are identical. }
  \label{tab:results}
\end{table*}

\timv{Consider making a table that summarizes this section: original scoring rule name | citation | mathematical definition | monotonic approximation. }\clara{liking this...}

\paragraph{Mutual Information.}\label{sec:mi} Maximum mutual information decoding \citep{li-etal-2016-diversity} aims to alleviate the inherent preference of neural models for high-frequency tokens when using the log-probability decoding objective. Rather than choosing the hypothesis $\yy$ to maximize conditional probability with respect to the input $\xx$, we instead choose $\yy$ to maximize pointwise mutual information (PMI):
\begin{equation}
    \mathrm{PMI}(\xx;\yy) = \log \frac{p( \xx, \yy)}{p(\xx)p(\yy)} \label{eq:pmi-def}
\end{equation}
Note that \cref{eq:pmi-def} is equivalent to $\log \frac{p(\yy \mid \xx)}{ p(\yy)}$, which can be rewritten as $\log p(\yy \mid \xx) - \log p(\yy)$ making the objective additive and thus \cref{eq:pmi-def} can conform to \cref{eq:additive}.

From this last form, we can see how mutual information decoding penalizes high-frequency and generic outputs; the negative $ p(\yy)$ term, as \citet{li-etal-2016-diversity} point out, acts as an ``anti-language model.'' One unfortunate side effect of this objective is that ungrammatical and nonsensical outputs, which have probabilities close to 0 under a language model like $ p(\yy)$, end up with high scores due to the second term in the score function. To address this problem, and to upper-bound the scoring function, we propose lower-bounding the language model term by a hyperparameter $1 \geq \eps > 0$. We additionally use the strength hyperparameter $\lambda$ employed by \citet{li-etal-2016-diversity}:
\begin{align}
    \score_{\textsc{pmi}}(\xx, \yy) = \,&\log p(\yy \mid \xx) \nonumber \\
     & \ \ \ - \lambda\log \max\{p(\yy), \eps\} \label{eq:pmi}
\end{align}

Similarly to our methods for length normalization, we can use the scoring function in \cref{eq:pmi} either with the heuristic:
\small
\begin{equation}
    h(\xx, \yy) = 
 \begin{cases}
    0 & \!\text{for } \yy.\mathrm{last}() = \eos \\ 
    - \lambda\log \eps (n_{max}\!\! - \!|\yy|)\! & \!\text{for } \yy.\mathrm{last}() \neq \eos
  \end{cases}
  \label{eq:mi_heur}
\end{equation}
\normalsize
\noindent or with stopping criterion as in \cref{eq:stop} albeit with $\score_{\textsc{pmi}}$ and upper-bound function:
\begin{equation}
    \mathcal{U}(\xx,\yy) =  - \lambda \log\eps (n_{max} - |\yy|)
\end{equation}

\noindent Since $-\lambda \log\eps$ is the best possible score at any given time step, clearly we can bound the increase in $\score_{\textsc{pmi}}$ by the above function. However, as with our length normalization strategy, we lose the $k$-optimality guarantee with the heuristic method for mutual information decoding.
We present experimental results for both methods in \cref{sec:exps}.

\section{Experiments}\label{sec:exps}
\begin{figure}[t!]
  \includegraphics[width=\linewidth]{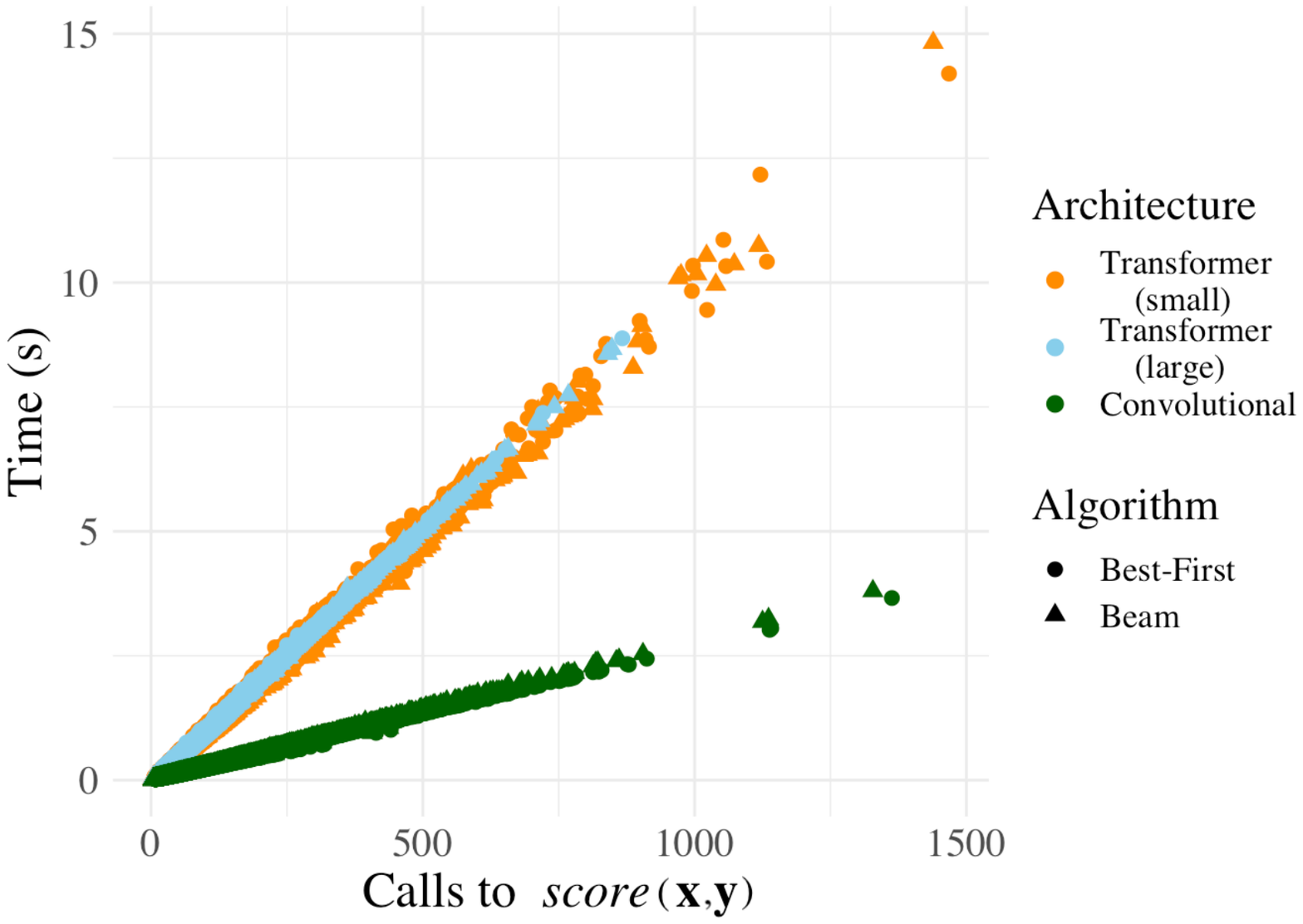}
  \caption{Number of calls to scoring function $\score$ vs. total sequence generation time. Each point is a decoded sequence. Colors represent different model architectures and shapes signify the decoding algorithm used (beam sizes 3 and 10 are included for each). There is no notable difference in the overhead (time-wise) of \algname{} and beam search. }
  \label{fig:time_v_pops}
\end{figure}

We run our algorithm on several language-related tasks that typically use beam search for decoding: neural machine translation (NMT) and abstractive summarization (AS). Specifically, experiments are performed on IWSLT'14 De-En \cite{IWSLTbib}, WMT'17 De-En \cite{bojar-EtAl:2017:WMT1}, MTTT Fr-En \cite{dataset2}, and CNN-DailyMail \cite{CNN_DM} using both Transformers \cite{vaswani2017attention} and Convolutional sequence-to-sequence models \cite{gehring-etal-2017-convolutional}.

For reproducibility, we use the data pre-processing scripts provided by fairseq \cite{ott2019fairseq} and follow their methods for training sequence transduction models. Hyperparameter are set in accordance with previous works. Specifically, on IWSLT'14 and MTTT tasks, we follow the recommended Transformer settings for IWSLT'14 in fairseq,\footnote{\url{https://github.com/pytorch/fairseq/tree/master/examples/translation}} which are based on \citet{vaswani2017attention} and \citet{gehring-etal-2017-convolutional}.
Hyperparameters for models trained on the WMT task are set following version 3 of the Tensor2Tensor toolkit~\cite{tensor2tensor}. 
We use byte-pair encoding (BPE; \citealt{sennrich2016bpeacl}) for all languages. Vocabulary sizes for WMT and IWSLT'14 are set from recommendations for the respective tasks in fairseq; for the MTTT tasks, vocabulary sizes are tuned on models trained with standard label-smoothing regularization. Similarly, the CNN/DailyMail dataset is pre-processed and uses BPE following the same steps as \cite{lewis2019bart}; model hyperparameters are likewise copied. Details are available on fairseq's website.\footnote{\url{https://github.com/pytorch/fairseq/blob/master/examples/bart/README.cnn.md}}\looseness=-1

We use \bleu \cite{Papineni:2002:BMA:1073083.1073135} (evaluated using SacreBLEU \cite{sacrebleu}) for MT metrics and \textsc{rouge-l} \cite{lin-2004-rouge} for abstractive summarization metrics.
We build our decoding framework in SGNMT.\footnote{\url{https://github.com/ucam-smt/sgnmt}}

\subsection{Running Time}
In \cref{tab:results}, we report values as the average number of calls to the scoring function per input; we do not use wall-clock time as this is heavily dependent on hardware. See \cref{fig:time_v_pops} for empirical justification of the correlation between calls to the scoring function and runtime on the hardware our experiments were run on. For reference, in our experiments, the scoring function took on average $>99\%$ of the total computation time, even with larger beam sizes, when overhead of the search algorithm is most significant.\looseness=-1

We find that best-first (BF) beam search leads to significant speed-ups over both traditional beam search and beam search with early stopping, with a performance increase\footnote{Performance increase is defined as $(\mathrm{old} - \mathrm{new})/ \mathrm{new}$} of $\approx 8x$ for a beam size of 500. We likewise find that \algname{} offers speed-ups over early stopping methods that are \emph{not} guaranteed to return the same results as standard beam search (see \cref{tab:stop_methods}). \looseness=-1

\begin{table}[!ht]
    \centering
    \small
    \setlength\tabcolsep{6pt}
    \begin{tabular}{lllll}
    \toprule
    \multicolumn{5}{c}{\bf IWSLT'14 De-En} \\
     $k$ & method & \makecell[c]{search\\error}&  \bleu &  \# calls \\
    \hline
    \multirow{3}{*}{\textbf{10}}& shrinking&0\% & 35.4 &229\dd{0}\rule{0pt}{3ex}\\

     & early&0\% & 35.4 &225\dd{2}\\
    & BF BS &-& 35.4 & \bf 169\dd{36}\\
    \hline
    \multirow{3}{*}{\textbf{100}}& shrinking& 31.7\% & 13.2 &2278\dd{0}\rule{0pt}{3ex}\\

     & early& 31.7\% & 13.2 &1738\dd{31}\\
    & BF BS &-& 34.7 &  \bf1275\dd{79}\\
     
     \midrule
      \multicolumn{5}{c}{\bf WMT'17 De-En} \\
      \multirow{3}{*}{\textbf{10}} &shrinking& 0\%&28.6 & 260\dd{0}\rule{0pt}{3ex}\\

    &early&0\%& 28.6 & 252\dd{3} \\
    &BF BS &-&28.6  &\bf 230\dd{12}\\
    \hline
    \multirow{3}{*}{\textbf{100}}& shrinking&1.7\% &26.4 &  2587\dd{0}\rule{0pt}{3ex}\\

     &early& 1.7\%&26.4 &2402\dd{8} \\
     &BF BS &- &26.9&\bf 2046\dd{26} \\
      
     \bottomrule
    \end{tabular}
    \caption{\bleu, search error, and average number of calls to $\score$ for different stopping criterion. ``shrinking'' refers to the shrinking beam method of \citet{bahdanau2014neural} and ``early'' refers to the stopping criterion of \citet{huang-etal-2017-finish}. Note that neither method is guaranteed to return the same result as standard beam search. Search error and performance increases are with respect to standard beam search.}
    \label{tab:stop_methods}
\end{table}

\subsection{Length Normalization}
We experiment with both forms of length normalization presented in \cref{sec:length} and provide results in \cref{tab:length_iwslt}. We find that both methods, i.e., changing the stopping criterion and using a heuristic during search, provide improvements over baseline \bleu scores albeit with different hyperparameter settings; increases are similar to improvements reported by \citet{murray-chiang-2018-correcting}. Notably, using a heuristic causes a large percentage of search errors with respect to standard beam search using the same scoring function. However, the difference in results appears to be \emph{beneficial} in terms of \bleu.\looseness=-1

\begin{table}[t]
    \centering
    \small
    \setlength\tabcolsep{4pt}
    \adjustbox{max width=\linewidth}{
    \begin{tabular}{lllllll}
    \toprule
\multicolumn{6}{c}{\bf IWSLT'14 De-En} \\
   & $k$ & $\beta$ & $b$ & \# calls & \makecell[c]{search\\error} & \bleu  \\
    \hline
    \multirow{2}{*}{\bf Heuristic} & 5 & 0.8 & $|\xx|$ &115\dd{0} & 40.6\% &33.9\db{0.3} \rule{0pt}{3ex} \\
     & 10 & 1.2 & $|\xx|$ & 229\dd{0}&54.7\% &33.8\db{0.5} \\
    \multirow{2}{*}{\bf \makecell{Stopping\\Criterion}} &5  & 0.5 & $\nmax$& 73\dd{58}&- &33.7\db{0.1} \\ 
    & 10 & 0.5 & $\nmax$& 130\dd{76}& - &33.7\db{0.4} \\
    \midrule
    
    \multicolumn{6}{c}{\bf MTTT Fr-En} \rule{0pt}{3ex}\\
    \multirow{2}{*}{\bf Heuristic} & 5  & 0.8 & $.7 |\xx|$ &100\dd{8}& 16.2\% &33.5\db{0.2} \rule{0pt}{3ex}\\
     & 10 & 1.0 & $.7 |\xx|$ & 196\dd{9} &25.2\% &33.6\db{0.6}   \\
    \multirow{2}{*}{\bf \makecell{Stopping\\Criterion}} & 5 &  1.0 & $\nmax$ & 65\dd{66}&-&34.1\db{0.8}   \\
     & 10 & 1.2 & $\nmax$& 88\dd{143}&- &34.1\db{1.1}  \\
     \bottomrule
    \end{tabular}}
    \caption{\bleu search error, and average number of calls to $\score$ for output obtained with length normalization scoring function on the IWSLT'14 De-En and MTTT Fr-En test sets. Increase in \bleu is over baseline with no length normalization. Search error and performance increases are with respect to standard beam search decoding using the same scoring function.}
    \label{tab:length_iwslt}
\end{table}

\subsection{Mutual Information}

We train a language model on the IWSLT dataset and use it to calculate $p(\yy)$ from \cref{eq:pmi} as marginalizing over $\yy$ is intractable (see \citet{li-etal-2016-diversity} for further justification). We run experiments using both of the methods discussed in \cref{sec:mi} and present results in \cref{tab:mi}. We find that both methods provide results of equivalent \bleu score compared with the baseline output, i.e., results obtained with the unbounded PMI objective and beam search. Again, despite the high search error rate demonstrated by the heuristic method, evaluation metrics are still comparable.

\subsection{Memory Usage}

We conduct a set of experiments where we limit total queue capacity to $k \cdot \gamma$ for $\gamma \in \{1, \dots, n_{max}\}$, as described in \cref{sec:imp}, and report the \bleu score of the resulting set of hypotheses.

As shown in \cref{tab:memory}, we find that restricting the queue capacity does not harm output quality and additionally, leads to even greater runtime performance increase. For example, runtime for decoding of IWSLT'14 with a beam size of 10 can be improved by $>\!3x$ while returning results with better evaluation metrics. We find that improvements are even more pronounced for larger beam sizes. Across beam widths and tasks, we find that search error (with respect to standard beam search) is quite low for $\gamma = 5$. Additionally, for  smaller $\gamma$, the change in \bleu score demonstrates that search error in this context does not necessarily hurt the quality of results.  
\begin{table}[t]
    \centering
    \small
    \setlength\tabcolsep{4pt}
    \begin{adjustbox}{width=\columnwidth}
    \begin{tabular}{lllllll}
    \toprule
   & $k$ & $\eps$\!&  $\beta$ & \# calls &   \makecell[c]{search\\error} & \bleu\\
    \hline
    \multirow{2}{*}{\bf Baseline} & 5 & - & .05 &115& -& 33.2  \rule{0pt}{3ex}\\
     &10 & - & .05 &229& -& 33.0  \\
    \multirow{2}{*}{\bf Heuristic}& 5  & .02 &.05 &129\dd{0}& 42.7\% & 33.2  \\
     &10 & .02 &.05&256\dd{0} &42.7\% &33.0  \\
    \multirow{2}{*}{\bf \makecell{Stopping\\Criterion}}& 5 & $3\mathrm{e}{\text{-}4}$ & .05 & 114\dd{1}& 29.2\% &33.2 \\
     &10 &  $5\mathrm{e}{\text{-}5}$ & .05 & 224\dd{2} & 26.6\% &33.0   \\
     \bottomrule
    \end{tabular}
    \end{adjustbox}
    \caption{\bleu scores with mutual information scoring function on IWSLT'14 De-En. Baseline is PMI decoding with \emph{unbounded} $p(\yy)$, i.e., $\eps = 0$. Search error is with respect to beam search decoding of baseline with same $\beta$. }
    \label{tab:mi}
\end{table}
\begin{table}[t]
    \centering
    \small
    \setlength\tabcolsep{6pt}
    \begin{tabular}{lllll}
    \toprule
    \multicolumn{5}{c}{\bf IWSLT'14 De-En} \\
    $k$ & $\gamma$  & \makecell[c]{search\\error} &   \bleu & \# calls\\
    \hline
    \multirow{3}{*}{\textbf{5}} & 2 &22.7\%  & 35.7\db{0.1} &43.8\dd{163}  \rule{0pt}{3ex}\\

    & 5 & 4.4 \% & 35.8\db{0.2}& 79.8\dd{44}\\
    & $\nmax$ & - & 35.6 & 93.0\dd{24}\\
    \hline
    \multirow{3}{*}{\textbf{10}}& 2 &22.6\% & 35.7\db{0.3} &48.4\dd{374}\rule{0pt}{3ex}\\

     & 5 &  4.5\%&35.6\db{0.2} &126.9\dd{81}\\
     & $\nmax$ & - & 35.4  & 169.0\dd{36} \\
     
     \midrule
      \multicolumn{5}{c}{\bf WMT'17 De-En} \\
      \multirow{3}{*}{\textbf{5}} & 2 &29.0\% &29.7\db{0.2} & 77.5\dd{75} \rule{0pt}{3ex}\\

    & 5 & 1.2\%& 29.5\db{0.0}& 115.8\dd{12} \\
    & $\nmax$  & -& 29.5 & 118.8\dd{10}\\
    \hline
    \multirow{3}{*}{\textbf{10}}& 2 &36.6\% &29.5\db{0.2} &97.3\dd{165}\rule{0pt}{3ex}\\

     & 5 &  2.6\% & 29.3\db{0.0} & 230.0\dd{12}\\
     & $\nmax$ & - & 29.3 & 230.2\dd{12}\\
      
     \bottomrule
    \end{tabular}
    \caption{\bleu scores and the number of calls to $\score$ on the IWSLT'14 De-En validation set and WMT'17 De-En test set with queue size restricted to $\nmax \cdot k$. Note that $\gamma\!=\!\nmax$ is the standard \algname{} algorithm. Performance increases are over standard beam search. Search error is with respect to beam search with same beam width.}
    \label{tab:memory}
\end{table}

\clara{I'd like to include a small section on search errors here. Just pointing to some research that they're not necessarily bad for language generation tasks}

\section{Related Work}

Our work is most similar to that of \citet{iterative_deepening}, who propose beam stack search. However, they are focused on exact inference and still evaluate hypotheses in breadth-first order. Additionally, their algorithm requires  $\calO(\nmax k)$ memory; while \algname{} has the same requirements, we introduce effective methods for reducing them, namely memory-reduced \algname.

\citet{huang-etal-2017-finish} propose and prove the optimality of an early-stopping criterion for beam search. The authors find in practice though that reduction in computation from their algorithm was generally not significant. We build on this work and introduce additional methods for avoiding unnecessary computation. Our method leads to better performance, as shown in \cref{tab:results}.

\citet{klein-manning-2003-parsing} use $\astar$ for PCFG parsing; however, they use the un-pruned version for exact search which is not applicable for NMT or AS as the memory requirements of the algorithm are far too large for these tasks. Subsequently, \citet{pauls-klein-2009-hierarchical} provide a method for pruning this search algorithm, albeit using a threshold rather than explicitly limiting the state space. \citet{huang-etal-2012-iterative} also adapt $\astar$ for a $k$-best decoding algorithm. While their methods differ notably from ours, they likewise employ pruning techniques that allow for substantial speedups.%

\citet{stahlberg-byrne-2019-nmt} create an exact inference algorithm for decoding and use it to analyze the output of neural NMT models. While they likewise employ the monotonicity of the scoring function to make their method tractable, they do not focus on speed or mimicking the results of standard beam search.

\section{Conclusion}
We propose \algname{}, an algorithm that allows for faster decoding while still guaranteeing $k$-optimality. We provide results on several sequence-to-sequence transduction tasks that show the speed-ups our algorithm provides over standard beam search for decoding neural models. We adapt several popular alternate scoring functions to \algname{} and provide a framework that can be used to adapt other scoring methods such as coverage normalization \cite{Wu2016GooglesNM} or diverse beam search \cite{diverse-beam-search}. We also provide a memory-reduced version of our algorithm, which returns competitive results in a fraction of the time needed for standard beam search.

\bibliography{anthology,acl2020}
\bibliographystyle{acl_natbib}

\newpage
\appendix

\end{document}